\documentclass[letterpaper, 10 pt, conference]{ieeeconf}  
\usepackage[T1]{fontenc}
\usepackage{graphicx}
\usepackage{hyperref}
\IEEEoverridecommandlockouts                              
\usepackage{siunitx} 
\usepackage{booktabs} 
   
\usepackage{amsmath}
\usepackage{amssymb}

\title{\LARGE \bf
GS-\text{I$^{3}$}: Gaussian Splatting for Surface Reconstruction from Illumination-Inconsistent Images
}

\author{Tengfei Wang\textsuperscript{†}, Xin Wang\textsuperscript{†}, \textit{Member, IEEE}, Yongmao Hou, Zhaoning Zhang, Yiwei Xu, \\and Zongqian Zhan, \textit{Member, IEEE}
\thanks{† Tengfei Wang and Xin Wang contributed equally to this work.}
\thanks{* This work was supported by the National Natural Science Foundation of China (No.42301507) and Natural Science Foundation of Hubei Province, China (No. 2022CFB727).}
\thanks{Tengfei Wang, Xin Wang, Yongmao Hou, Zhaoning Zhang, Yiwei Xu, and Zongqian Zhan are with the School of Geodesy and Geomatics, Wuhan University, China PR. \textit{ Corresponding Author: Zongqian Zhan, zqzhan@sgg.whu.edu.cn})
}%
}

\begin{document}

\maketitle
\thispagestyle{empty}
\pagestyle{empty}

\begin{figure*}[htbp]
    \centering
    \includegraphics[width=0.8\linewidth]{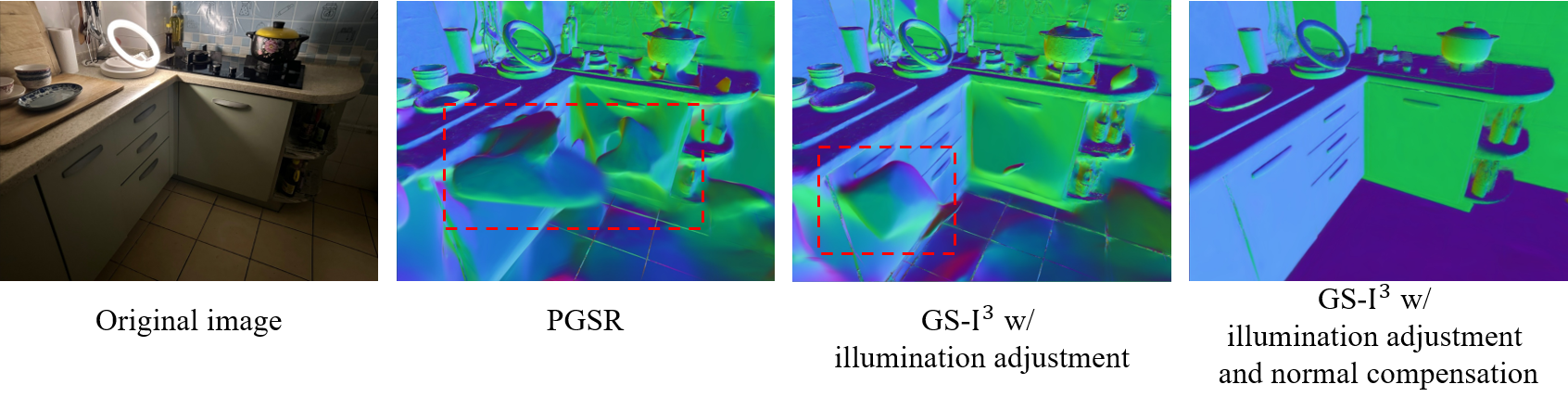}
    \caption{Original image on the Kitchen data(subset of Gaussian in the dark dataset), along with the normal maps obtained by the different methods.}
    \label{fig:show}
\end{figure*}

\begin{abstract}

Accurate geometric surface reconstruction, providing essential environmental information for navigation and manipulation tasks, is critical for enabling robotic self-exploration and interaction. Recently, 3D Gaussian Splatting (3DGS) has gained significant attention in the field of surface reconstruction due to its impressive geometric quality and computational efficiency. While recent relevant advancements in novel view synthesis under inconsistent illumination using 3DGS have shown promise, the challenge of robust surface reconstruction under such conditions is still being explored. To address this challenge, we propose a method called GS-\text{I$^{3}$}. Specifically, to mitigate 3D Gaussian optimization bias caused by underexposed regions in single-view images, based on Convolutional Neural Network (CNN), an adaptive illumination adjustment framework is introduced. Furthermore, inconsistent lighting across multi-view images, resulting from variations in camera settings and complex scene illumination, often leads to geometric constraint mismatches and deviations in the reconstructed surface. To overcome this, we propose a normal compensation mechanism that integrates reference normals extracted from single-view images with normals computed from multi-view observations to effectively constrain geometric inconsistencies. Extensive experimental evaluations demonstrate that GS-\text{I$^{3}$} can achieve robust and accurate surface reconstruction across complex illumination scenarios, highlighting its effectiveness and versatility in this critical challenge. The code is available: \href{ https://tfwang-9527.github.io/GS-3I/}{https://tfwang-9527.github.io/GS-3I/}
\end{abstract}


\section{INTRODUCTION}
Accurate geometric surface reconstruction is a fundamental support for robotic self-exploration and interaction, providing crucial 3D environmental information for tasks such as navigation, manipulation, decision-making, etc. \cite{robotic_1,robotic_2}. As robots increasingly operate in unstructured and complex environments, the ability to perform surface reconstruction with high precision, robustness, and adaptability becomes increasingly critical. Traditional methods for surface reconstruction often rely on multi-view stereo (MVS) techniques or depth sensors \cite{mvs_1,mvs_2}, which, while effective in controlled conditions, typically involve complex computational steps and are time-consuming, thus limiting their applicability in practice.

In contrast, recent advancements in 3D Gaussian Splatting (3DGS) \cite{3DGS} have introduced a promising approach to surface reconstruction. Until now, 3DGS has attracted significant attention due to its superior geometric quality and computational efficiency, positioning it as a promising candidate for real-time applications in challenging environments \cite{sugar,2dgs}. Its capability to produce high-quality surface representations marks a notable improvement over traditional techniques \cite{colmap}.

However, despite these advances, 3DGS still faces significant challenges when dealing with inconsistent illumination, a common issue in real-world scenarios \cite{vastgaussian,darkgs}. Although 3DGS has demonstrated considerable success in novel view synthesis under such conditions \cite{DK_gaussian,hdr-gs,hdrsplat}, the task of achieving robust and accurate surface reconstruction remains unresolved. Inconsistent lighting can introduce 3D Gaussian optimization biases, disrupt geometric constraints, and lead to deviations in the reconstructed surface. These challenges ultimately compromise the accuracy, fidelity, and reliability of the surface reconstruction, hindering its practical application in complex environments where lighting conditions are highly variable.

To address these challenges, we propose a novel method called GS-\text{I$^{3}$}. Our approach is designed to mitigate the adverse effects of inconsistent illumination on surface reconstruction by introducing two key innovations: an adaptive illumination adjustment framework based on CNN and a normal compensation mechanism that integrates reference normals extracted from single-view images with normals computed from multi-view observations.

The first contribution, the adaptive illumination adjustment framework, is designed to address the optimization bias caused by underexposed regions in single-view images. Underexposed regions often contain insufficient texture and detail, leading to inaccurate depth estimates and surface reconstructions as shown in Fig.\ref{fig:show}. By leveraging a CNN-based adaptive illumination adjustment framework, our method can adaptively adjust the exposure levels of these regions, ensuring that the optimization process is not biased towards overexposed or underexposed areas. This correction framework enhances the overall quality of the reconstructed surface by providing more reliable images for the reconstruction process.

The second contribution, the normal compensation mechanism, tackles the problem of geometric constraint mismatches caused by inconsistencies in lighting across multi-view images. Variations in camera settings and complex scene illumination can lead to significant differences in the appearance of the same scene across different views, resulting in mismatches in the geometric constraints used for reconstruction, as shown in Fig.\ref{fig:show}. To overcome this, our method integrates reference normals extracted from single-view images with normals computed from multi-view observations. This integration allows for a more robust and accurate estimation of surface normals, effectively constraining geometric inconsistencies and improving the overall quality of the reconstructed surface.

Extensive experimental evaluations demonstrate the efficacy of our GS-\text{I$^{3}$} in achieving robust and accurate surface reconstruction across complex illumination scenarios. Our method outperforms existing approaches in terms of geometric accuracy, highlighting its potential for real-world applications in robotic exploration and interaction.

\section{RELATED WORK}

In this section, two related topics are reviewed including surface reconstruction and inconsistent illumination processing

\subsection{Surface Reconstruction}

Traditional surface reconstruction methods have largely relied on multi-view stereo (MVS) pipelines, which triangulate correspondences across images to infer depth and geometry. These approaches, such as COLMAP \cite{colmap} and PMVS \cite{PMVS}, excel in structured environments with consistent lighting but struggle with textureless regions, occlusions, and computational inefficiency in large-scale scenes. To address these limitations, sensor fusion techniques integrating LiDAR, RGB-D cameras, or structured light have been proposed, offering higher accuracy at the cost of hardware complexity and limited scalability \cite{lidar_mvs}.

The advent of neural implicit representations, particularly Neural Radiance Fields (NeRF)\cite{nerf}, revolutionized the field by enabling high-fidelity scene modeling through differentiable volume rendering . Subsequent works extended NeRF for surface reconstruction by incorporating signed distance functions (SDFs) or occupancy networks, achieving impressive results in both geometry and appearance \cite{neus,volsdf}. However, these methods remain computationally intensive and sensitive to illumination variations due to their reliance on photometric consistency.

In recent years, 3D Gaussian Splatting (3DGS) has emerged as a groundbreaking approach by integrating explicit geometric primitives with differentiable rasterization techniques, enabling real-time rendering with remarkable efficiency. This method represents scenes using anisotropic 3D Gaussians, achieving state-of-the-art performance in both reconstruction quality and computational efficiency \cite{2dgs,pgsr}, making it particularly suitable for real-time interactive applications such as robotics. DN-Splatter \cite{DN-Splatter} attempts to enhance reconstruction accuracy using single-view normals, however, accurate normal prediction typically requires uniformly bright scene illumination. Extensions of 3DGS, including dynamic scene surface reconstruction \cite{dynasurfgs} and semantic-aware surface reconstruction \cite{pg-sag}, further demonstrate the versatility of this framework. While 3DGS has shown significant potential in the field of surface reconstruction, existing methods have limited exploration in addressing surface reconstruction under challenging conditions with significant illumination variations and inconsistent lighting. This research gap motivates our work to adapt 3DGS for illumination-robust geometric modeling, aiming to tackle the challenges posed by complex lighting environments.

\subsection{Inconsistent Illumination Processing}
Inconsistent lighting poses a fundamental challenge to multi-view reconstruction as it violates the brightness constancy assumption that most multi-view methods rely on \cite{traditinal-light1}. In the realm of NeRF-based 3D reconstruction methods, researchers have proposed various solutions. HDR-NeRF \cite{hdr-nerf} and Raw-NeRF \cite{raw-nerf} have successfully recovered normally lit scenes from inconsistently lit images by utilizing HDR and RAW format data. However, the stringent requirements for input data make these methods difficult to apply directly to RGB data commonly captured by robots. NeRF-W \cite{nerf-w} attempts to directly process RGB images in wild environments by modeling lighting variations as appearance embeddings, which to some extent mitigates the artifacts in rendering. Nevertheless, the lengthy computation time of this method severely limits its application in robotic exploration tasks that require high real-time performance.

In recent years, 3DGS technology has demonstrated significant advantages in the field of 3D reconstruction due to its fast and efficient rendering capabilities. In response to the rendering challenges under inconsistent lighting conditions, new methods based on 3DGS have emerged. Vastgaussian \cite{vastgaussian} decouples the appearance of scenes under complex lighting and applies mapping transformations to rendered images to adapt to appearance changes; LO-Gaussian \cite{LO_gaussian} innovatively introduces the concept of a simulated filter between real and rendered images, recovering scenes from overexposed and overly dark lighting by simulating the degradation process; Dark in the Gaussian \cite{DK_gaussian} approaches from the perspective of exposure, designing a camera response module to compensate for lighting inconsistencies in multi-view scenarios. WildGaussians \cite{wildgaussians} addresses dynamic outdoor environments through a dual-embedding mechanism that separately models per-image global illumination and per-Gaussian local lighting effects, achieving robust reconstruction under varying weather conditions. Meanwhile, GS-W \cite{gsw} proposes an adaptive appearance modeling framework that combines static material properties with dynamic environmental features through a learnable fusion network, enabling precise control over scene appearance under inconsistent illumination. However, these methods primarily optimize for rendering tasks, and a systematic solution for surface reconstruction in inconsistently lit scenes remains lacking, which is a critical issue that current research urgently needs to address.

\section{Method}
\begin{figure*}[htbp]
    \centering
    \includegraphics[width=1.0\linewidth]{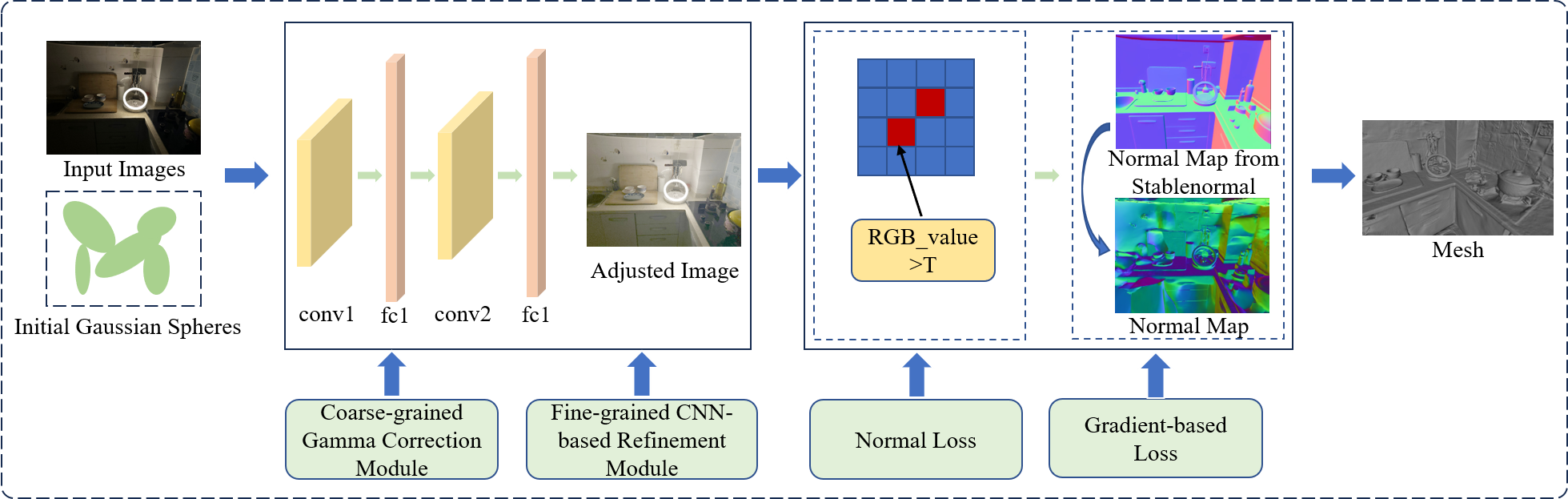}
    \caption{The workflow of GS-\text{I$^{3}$}.}
    \label{fig:overflow}
\end{figure*}

In the task of surface reconstruction under inconsistent illumination conditions, two critical challenges are predominantly encountered: 3D Gaussian optimization bias induced by exposure variation within a single view, and inconsistency in Geometry caused by illumination variations across multiple views. To address these limitations, this study proposes a novel surface reconstruction method, termed GS-\text{I$^{3}$}, specifically designed for scenes with inconsistent illumination. The proposed method comprises two pivotal components: Firstly, in Section \ref{sec:Illumination Adjustment}, we introduce an adaptive illumination adjustment framework based on CNN, which effectively balances illumination inconsistencies within a single view. Secondly, in Section \ref{sec:Normal Constraint}, we devise a normal compensation mechanism that utilizes pre-trained normal maps from single views to rectify reconstruction errors arising from illumination-inconsistent images. The workflow is shown in Fig.\ref{fig:overflow}.

\subsection{Adaptive Illumination Adjustment Framework}
\label{sec:Illumination Adjustment}

The adaptive illumination adjustment framework is designed to achieve intelligent dynamic range optimization via a novel two-stage modulation mechanism, integrating a coarse-grained Gamma correction module and a fine-grained CNN-based refinement module. The proposed framework follows a global-local optimization paradigm, where pixel-wise Gamma parameters are adaptively derived from their relative luminance positions in the cumulative distribution function (CDF), modulated by learnable weights to ensure initial contrast enhancement. A lightweight convolutional network then performs local illumination refinement, with spatially adaptive parameter fusion enabling precise dynamic range adjustment.

\begin{itemize}
\item \textbf{Coarse-grained Gamma Correction Module}
\end{itemize}

In the initial stage of scene visual information processing, a coarse-grained Gamma correction modulation based on pixel spatial distribution is applied to all input images. The core of this process lies in learning, for each image, its own dynamically adjustable Gamma value feasible range parameters \( [\gamma_{\text{min}}, \gamma_{\text{max}}] \). For any pixel in the image, its corresponding Gamma value is not globally uniform but is dynamically computed based on the relative position of that pixel's brightness value within the overall image brightness distribution.  

Specifically, this computation is mapped as a linear function: the percentile rank of the pixel's brightness in the image's cumulative distribution function (CDF) serves as the input, and a linear interpolation maps it to the predefined Gamma value interval \( [\gamma_{\text{min}}, \gamma_{\text{max}}] \) of the image. This adaptive mechanism, based on the relative rank (Rank) of pixel brightness, enables different brightness regions to be assigned differential nonlinear enhancement strengths during the coarse processing stage. The goal is to preliminarily optimize the contrast distribution of the image, enhance local detail visibility, and lay the foundation for subsequent global optimization.

Here, the Gamma values $\gamma_{\text{min}}$ and $\gamma_{\text{max}}$ are derived as
\begin{equation}
\begin{cases}
\gamma_{\text{min}} = a \cdot \exp(b), \\
\gamma_{\text{max}} = c \cdot \exp(d),
\end{cases}
\label{eq:gamma_minmax}
\end{equation}
where $a, b, c, d$ denote learnable model parameters optimized during training. For a given pixel with brightness percentile rank $p$ ($p \in [0, 1]$), the linear interpolation of the Gamma correction is defined as
\begin{equation}
\gamma(p) = \gamma_{\text{min}} + p \cdot (\gamma_{\text{max}} - \gamma_{\text{min}}).
\label{eq:gamma_p}
\end{equation}

This formulation ensures that Darker pixels (with lower $p$) receive stronger enhancement, significantly increasing their brightness to reveal details in underexposed areas. Meanwhile, brighter pixels (with higher $p$) undergo moderate suppression, effectively reducing their intensity to prevent overexposure while maintaining highlight details, thereby achieving adaptive contrast enhancement through context-aware parameterization. The entire image is processed to obtain an adaptively gamma-mapped image $I_{\text{gm}}$.
\begin{equation}
I_{\text{gm}}(x, y) = [I_{\text{gt}}(x, y)]^{\gamma(p(x, y))}.
\label{eq:I_gm}
\end{equation}

\begin{itemize}
\item \textbf{Fine-grained CNN-based Refinement Module}
\end{itemize}

Building upon coarse-grained gamma correction, the framework incorporates a lightweight convolutional network for fine-grained illumination modeling: Two convolutional layers are used to perform feature extraction on each image, yielding a \(224 \times 224\) illumination weight feature map. The specific parameters of the convolutions are as follows:  
\[F_1 = \operatorname{ReLU}\left( \text{Conv}_{\text{3}\times\text{3}\to64}(I_{\text{rendered}}) \right)\]
where the first convolution employs a \(3 \times 3\) kernel with 64 output channels, operating on the rendered image \(I_{\text{rendered}}\), followed by a ReLU activation function.  
\begin{equation}
F_2 = \operatorname{ReLU}\left( \text{Conv}_{\text{3}\times\text{3}\to1}(F_1) \right)
\label{eq:F_2}
\end{equation}
where the second convolution uses a \(3 \times 3\) kernel with 1 output channel, processing the feature map \(\mathcal{F}_1\) from the first layer, followed by a ReLU activation function.  

The feature map $F_2$ generated during feature extraction is physically fused with a learnable scene illumination map $M \in \mathbb{R}^{224\times224}$:
\begin{equation}
F_{\text{map}} = M \odot F_2
\label{eq:F_map}
\end{equation}

The fused feature map $F_{\text{map}}$ is bilinearly upsampled to match the rendered image's resolution, then combined and optimized it to generate $I_{\text{map}}$:

\begin{equation}
L_{\text{illum}} = \text{resize}(F_{\text{map}}) \odot I_{\text{rendered}}
\label{eq:map}
\end{equation}

This process achieves precise modeling of complex illumination distributions through collaborative optimization of spatial illumination parameters and local feature responses. The final illumination adjustment loss $L_{\text{illum}}$ is calculated as a weighted combination of SSIM and L1 losses between $I_{\text{map}}$ and $I_{\text{gm}}$, with \( \lambda \) (typically 0.2) as the balancing hyperparameter.
\begin{equation}
{L_{\text{illum}}}=(1-\lambda){L1}(I_{\text{map}},I_{\text{gm}})+\lambda{SSIM}(I_{\text{map}},I_{\text{gm}})
\label{eq:illum_loss}
\end{equation}

\subsection{Normal Compensation}
\label{sec:Normal Constraint}

The adaptive illumination adjustment framework can map the brightness of a single view to an appropriate range. However, illumination inconsistencies between different views can lead to significant fluctuations in Gaussian spheres, thereby reducing the accuracy of surface reconstruction. Building on the single-view adaptive illumination adjustment method, we devise a normal compensation mechanism that utilizes pre-trained normal maps from single views to rectify reconstruction errors arising from illumination-inconsistent images.

First, during the training process, we identify pixels requiring correction using a combined RGB loss \( L_{\text{illum}} \). 

We set a threshold \( T \) to determine whether a pixel requires correction. For pixels with \( L_{\text{illum}}(i) > T \), we compute a normal loss based on the difference between the predicted normals and the normals generated by a pre-trained large model, such as StableNormal \cite{stablenormal}. Trained on a large dataset, StableNormal provides robust normal predictions that capture fine geometric details. This normal loss is used to guide the correction of the predicted normals during training.

For pixels with \( L_{\text{illum}}(i) > T \), the normal loss is defined as the L1 difference between the predicted normals and the normals generated by the StableNormal model. The normal loss for the \( i \)-th pixel is given by:

\begin{equation}
L_{\text{normal}}(i) = \begin{cases}
0 & \text{if } L_{\text{illum}}(i) \leq T, \\
\| N_i^{\text{pred}} - N_i^{\text{StableNormal}} \|_1 & \text{if } L_{\text{illum}}(i) > T,
\end{cases}
\label{eq:normal_loss}
\end{equation}

where \( N_i^{\text{pred}} \) is the originally predicted normal, and \( N_i^{\text{SN}} \) is the normal predicted by StableNormal.

However, merely correcting the normals is insufficient for accurate surface reconstruction, as it only adjusts the normal directions without ensuring local consistency. To address this, we introduce a gradient-based loss term \( L_{\text{gradient}} \) that penalizes discrepancies between the gradients of the predicted normals and the gradients of the normals generated by StableNormal. This term encourages smooth and locally consistent normal fields, which are essential for high-quality surface reconstruction:

\begin{equation}
L_{\text{gradient}} = \sum_{i=1}^{n} \left\| \nabla N_i^{\text{pred}} - \nabla N_i^{\text{SN}} \right\|_1,
\label{eq:gradient_loss}
\end{equation}

where \( \nabla N_i^{\text{pred}} \) and \( \nabla N_i^{\text{SN}} \) are the spatial gradients of the predicted normals and the normals generated by StableNormal, respectively. This term ensures that the predicted normals exhibit smooth and locally consistent variations, which are critical for accurate surface reconstruction.

The final loss function combines  illumination adjustment loss \( L_{\text{illum}} \), normal loss \( L_{\text{normal}} \), gradient-based loss \( L_{\text{gradient}} \), and multi-view consistency Loss \( L_{\text{mvs}} \)  from PGSR \cite{pgsr} is expressed as:

\begin{equation}
L_{\text{total}} = \lambda_1 L_{\text{illum}} + \lambda_2 L_{\text{normal}} + \lambda_3 L_{\text{gradient}} + \lambda_4 L_{\text{mvs}}
\label{eq:total_loss}
\end{equation}

where $\lambda_1 = {1.0}$, $\lambda_2 = {0.15}$, $\lambda_3 ={0.0015}$, and $\lambda_4 ={0.03}$ are hyperparameters used to balance the weights of different loss terms.

\section{Experiments}
\subsection{Experimental Setup}
\begin{figure}[htbp]
    \centering
    \includegraphics[width=0.9\linewidth]{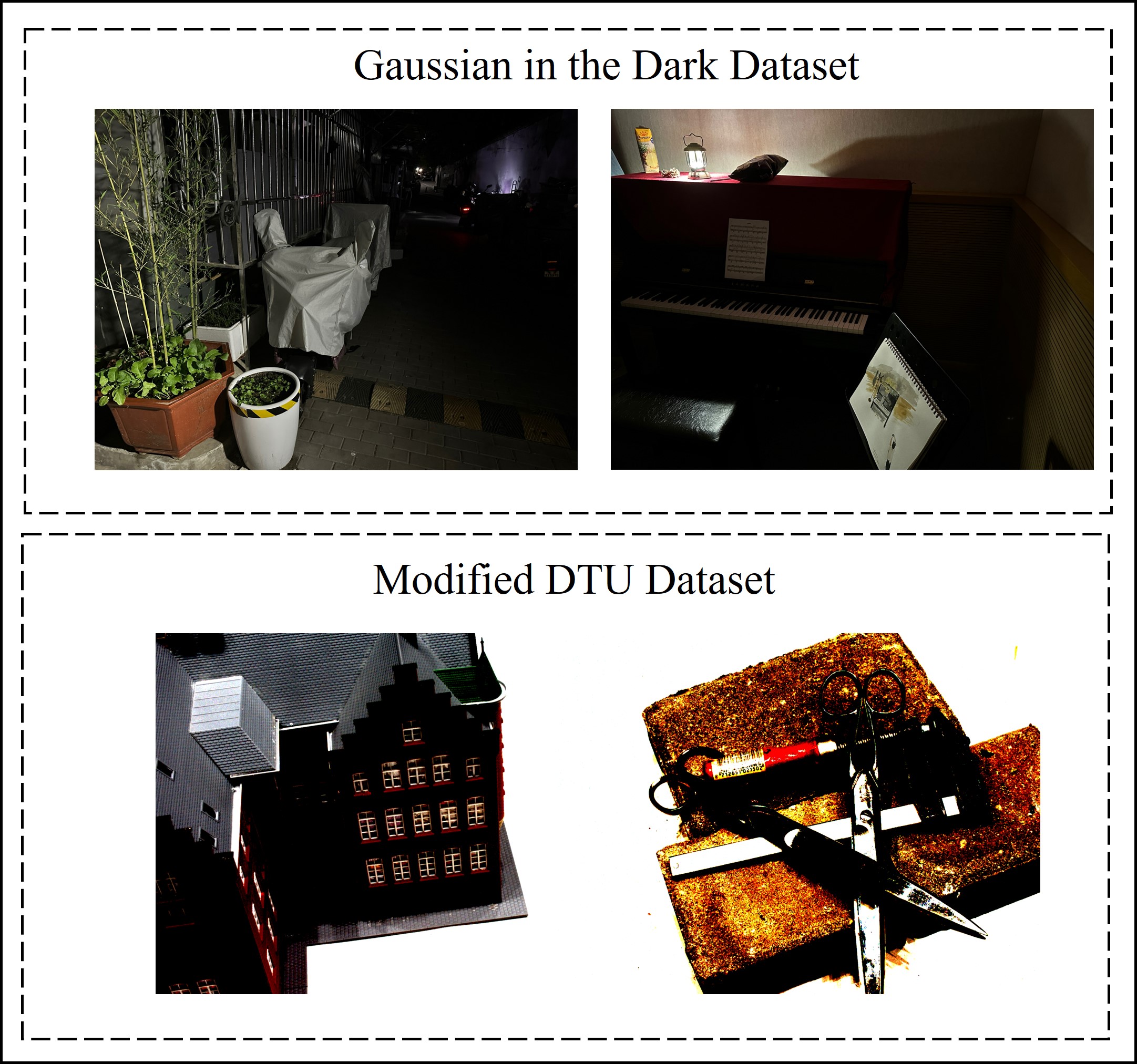}
    \caption{Sample images of  Gaussian in the Dark dataset and Modified DTU dataset.}
    \label{fig:dataset}
\end{figure}

\textbf{Datasets and Metrics}
To validate the effectiveness of our proposed method, GS-\text{I$^{3}$},  in scenarios with inconsistent illumination, we utilized the illumination-inconsistent dataset introduced by Gaussian in the Dark \cite{DK_gaussian}. This dataset comprises 12 real-world scenes (5 indoor and 7 outdoor), each containing approximately 80 to 130 naturally exposed images captured from multiple angles, with a resolution of 3991 × 2960. While this dataset is suitable for qualitative evaluation, it lacks ground truth for quantitative assessment. To address this limitation, we introduced a modified version of the DTU dataset \cite{dtu}, a widely used benchmark with ground truth, by adding random lighting perturbations to simulate inconsistent illumination conditions. Specifically, we applied random brightness and contrast adjustments (with factors between 0.5-1.5) followed by gamma corrections with randomly selected target gamma values of either 0.1 or 0.8 to create abrupt and varied illumination effects. This modified DTU dataset enables qualitative evaluation of our method. Sample images are illustrated in the accompanying Fig.\ref{fig:dataset}.

\textbf{Baselines and Implementation.}
To the best of our knowledge, our method is the first to tackle surface reconstruction using illumination-inconsistent images, there are no directly comparable baseline methods available. To evaluate the effectiveness of GS-\text{I$^{3}$}, we compared it with three state-of-the-art surface reconstruction methods: PGSR \cite{pgsr}, 2DGS \cite{2dgs}, GOF \cite{gof} and SuGaR \cite{sugar}, which are the most relevant ones to our work.

\subsection{Comparison with SOTA Methods}

\begin{figure*}[htbp]
    \centering
    \includegraphics[width=0.6\linewidth]{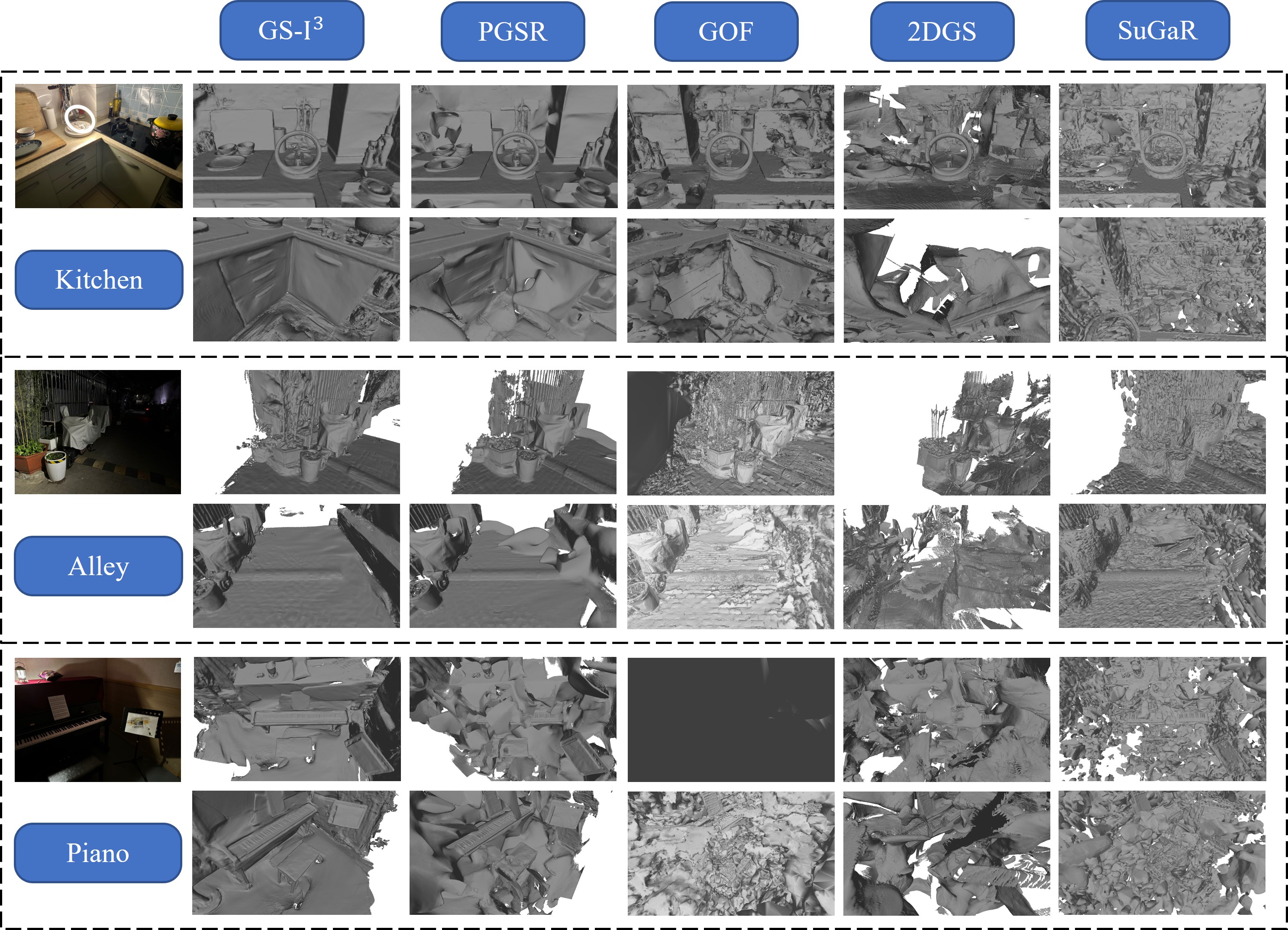}
    \caption{Comparison of mesh reconstruction results across the Gaussian in the Dark dataset with various methods.}
    \label{fig:dark_result}
\end{figure*}

\begin{figure*}[htbp]
    \centering
    \includegraphics[width=0.6\linewidth]{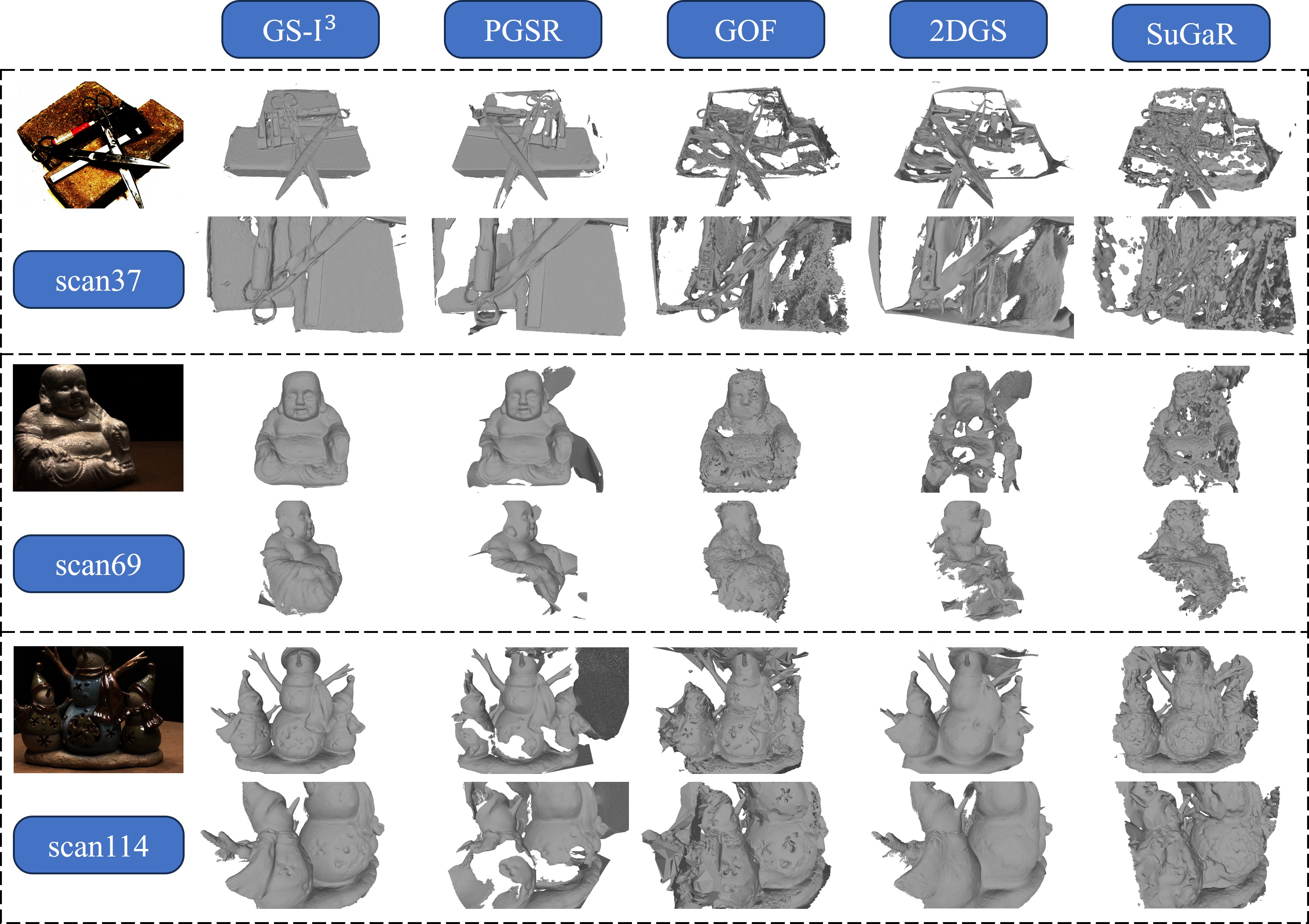}
    \caption{Comparison of mesh reconstruction results across the modified DTU dataset with various methods.}
    \label{fig:DTU_result}
\end{figure*}

Fig.\ref{fig:dark_result} and Fig.\ref{fig:DTU_result} present a qualitative comparison of the proposed GS-\text{I$^{3}$} method with other existing methods (PGSR, 2DGS, GOF, and SuGaR) in scenarios with inconsistent lighting conditions on modified DTU dataset and Dark in the Gaussian dataset. It is clearly observed that GS-\text{I$^{3}$} generates surface models that more accurately reflect the actual scene, whereas other methods exhibit significant reconstruction errors. These errors mainly manifest as surface geometry distortion, holes, and unrealistic geometric structures.

The primary cause of these errors lies in the inconsistent lighting across images captured from different angles in such scenes. The varying exposure results in considerable brightness differences for the same scene area across different images. These brightness discrepancies make it challenging for previous gaussian-based methods based on multi-view geometric constraints to accurately fit the true surface geometry during the training process. Specifically, conventional 3DGS-based methods rely heavily on gaussian splatting distribution to model the scene surface. However, due to lighting inconsistencies, Gaussian splatting attempts to satisfy rendering requirements from different viewpoints during optimization, leading to erroneous distributions in the air. While these erroneous distributions can reduce discrepancies between viewpoints during rendering, they significantly deviate from the actual surface geometry, causing a notable decline in reconstruction quality.

In contrast, GS-\text{I$^{3}$} addresses this issue with two key contributions. First, at the single-view level, GS-\text{I$^{3}$} introduces an adaptive illumination adjustment framework based on CNN, effectively balancing the lighting inconsistencies within a single image. Specifically, the method enhances the contrast of dark regions while preserving the details of bright areas, thereby strengthening the loss constraint in dark regions during training, which significantly improves the reconstruction quality of dark areas. Second, at the multi-view level, GS-\text{I$^{3}$} employs a normal compensation mechanism, leveraging pre-trained single-view normals to correct lighting inconsistencies across multiple views. This mechanism introduces pre-trained normal information, eliminating the lighting ambiguity for the same region across different views, thus providing a more accurate representation of the surface geometry.

The comparison results in Fig.\ref{fig:dark_result} and \ref{fig:DTU_result} demonstrate that GS-\text{I$^{3}$} is capable of reconstructing surface models with rich details and accurate geometry under complex lighting conditions, while other methods show noticeable distortions and errors in highlights, shadows, and dark areas. This outcome strongly validates the superiority of GS-\text{I$^{3}$} in scenes with inconsistent lighting.

Tab.\ref{tab:comparison} presents the quantitative comparison of the proposed method with other methods on the modified DTU dataset. The experiments cover multiple scenes, each involving varying degrees of lighting changes. The Chamfer Distance metric is employed for geometric accuracy evaluation, as it comprehensively reflects the geometric deviation between the reconstructed surface and the true surface. To further validate the effectiveness of our method, we performed experiments on the DTU dataset using traditional illumination correction techniques, including gamma transformation and histogram transformation, for image enhancement. These enhanced images were subsequently processed using both PGSR and GOF methods to evaluate the performance of traditional illumination processing in scene reconstruction.

As shown in the results in Tab.\ref{tab:comparison}, GS-\text{I$^{3}$} outperforms all other methods in every scene of the DTU dataset, achieving significantly lower chamfer distance values. This result is consistent with the qualitative analysis in \ref{fig:DTU_result} and further confirms the effectiveness of GS-\text{I$^{3}$} in scenarios with inconsistent lighting. Notably, while traditional illumination enhancement methods can improve reconstruction results to some extent, they require offline preprocessing, whereas our approach is end-to-end and achieves higher accuracy. Additionally, due to the low accuracy of mesh obtained by some methods, which makes it impossible to complete ICP(Iterative Closest Point) registration with the ground truth point cloud for evaluation of chamfer distance, we aligned the ground truth point cloud to the mesh obtained by COLMAP \cite{colmap} under original lighting conditions to complete the accuracy assessment. This approach ensures a fair and consistent evaluation across all methods.

\begin{table*}[htbp]
\centering
\caption{Comparison of different methods on the modified DTU dataset. Where the first line is the number of different data in the dataset, and the chamfer distance unit is centimeters(lower values indicate better performance, NAN means out of memory).}
\label{tab:comparison}
\begin{tabular}{l*{15}{S[table-format=1.2]}S[table-format=1.2]}
\toprule
\textbf{Method} & \textbf{24} & \textbf{37} & \textbf{40} & \textbf{55} & \textbf{63} & \textbf{65} & \textbf{69} & \textbf{83} & \textbf{97} & \textbf{105} & \textbf{106} & \textbf{110} & \textbf{114} & \textbf{118} & \textbf{122} & \textbf{Mean} \\ \midrule
SuGaR          & 4.14 & 2.96 & 1.66 & 1.29 & 4.63 & 2.97 & 1.99 & 3.83 & 2.14 & 3.63 & 2.88 & 1.86 & 3.19 & 4.92 & 2.58 & 2.98 \\ 
2DGS           & 3.71 & 3.24 & 2.01 & 1.76 & 3.70 & 2.34 & 2.40 & 4.11 & 3.66 & 2.40 & 2.80 & 3.46 & 2.88 & 3.08 & 4.29 & 2.90  \\ 
GOF            & 4.45 & 3.01 & 2.43 & 2.51 & 3.60 & 1.52 & 2.61 & 3.21 & 0.88 & 1.70 & 2.39 & 2.33 & 1.30 & 1.77 & 1.47 & 2.35 \\ 
GOF+gamma      & 4.21 & 3.22 & 2.12 & 2.43 & 3.88 & 1.86 & 2.22 & 3.22 & 2.61 & 2.41 & 2.91 & 1.93 & 2.36 & 2.39 & 1.61 & 2.63 \\ 
GOF+historgram & 4.51 & 2.65 & 2.31 & 2.90 & 3.95 & 2.56 & NAN & 3.42 & 2.58 & 2.15 & 2.88 & NAN & 3.25 & 2.92 & 2.12 & 2.94 \\ 
PGSR           & 3.09 & 1.50 & 2.05 & 1.13 & 2.02 & 0.98 & 2.74 & 3.14 & 1.65 & 0.84 & 1.79 & 1.57 & 1.16 & 2.07 & 0.71  & 1.76  \\ 
PGSR+gamma     & 2.98 & 0.86 & 1.01 & 0.48 & 1.77 & 1.13 & 1.40 & 2.25 & 0.74 & 1.03 & 0.87 & 1.02 & 1.40 & 1.20 & 0.53  & 1.24  \\ 
PGSR+historgram& 3.03 & 0.80 & 1.10 & 0.53 & 2.35 & 0.87 & 1.09 & 2.14 & 0.78 & 0.90 & 0.49 & 1.09 & 0.60 & 1.46 & 0.39  & 1.17  \\ 
\textbf{GS-\text{I$^{3}$}} & \textbf{2.81} & \textbf{0.74} & \textbf{0.91} & \textbf{0.39} & \textbf{1.72} & \textbf{0.78} & \textbf{0.63} & \textbf{2.03} & \textbf{0.61} & \textbf{0.65} & \textbf{0.47} & \textbf{0.88} & \textbf{0.45} & \textbf{1.09} & \textbf{0.36} & \textbf{0.97} \\ 
\bottomrule
\end{tabular}
\end{table*}

\subsection{Ablation Study}
\begin{figure}[htbp] 
    \centering
    \includegraphics[width=1.0\linewidth]{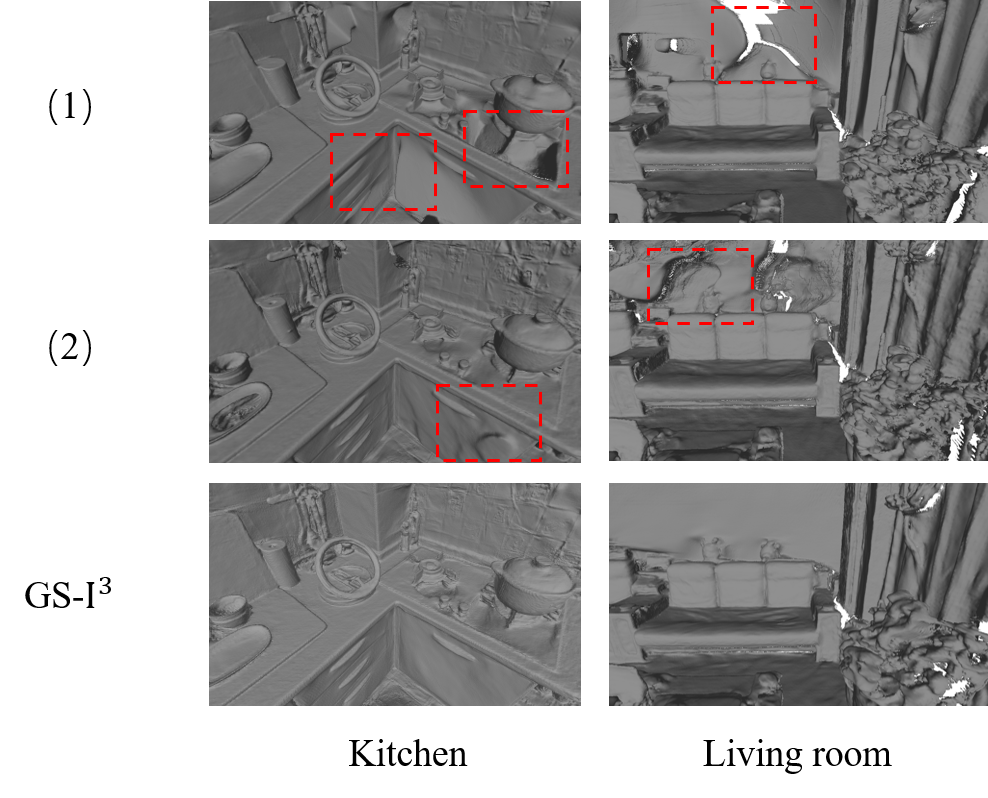}
    \caption{Comparison of different methods in the ablation study on the modified Gaussian in the dark dataset (selected scenes).}
    \label{fig:ablation_fig}
\end{figure}

To validate the effectiveness of the individual components in the GS-\text{I$^{3}$} method, we conducted ablation experiments on the modified DTU dataset, focusing on scan69 and scan105, as well as the Gaussian in the Dark dataset, focusing on the Kitchen and Living room subset. These experiments were designed to evaluate the contribution of each individual component to the overall performance. Specifically, we performed the following experiments: (1) using only the adaptive illumination adjustment framework, (2) using only the normal compensation. The qualitative results of these ablation experiments are shown in Fig.\ref{fig:ablation_fig}, while the quantitative results are presented in Tab.\ref{tab:dtu_ablation}.

\begin{table}[htbp]
\centering
\caption{Comparison of different methods in the ablation study on the modified DTU dataset (selected scans).}
\label{tab:dtu_ablation}
\setlength{\tabcolsep}{20pt} 
\begin{tabular}{c*{2}{S[table-format=1.2]}} 
\toprule
\textbf{Method} & \textbf{scan69} & \textbf{scan105} \\ \midrule
(1)         & 0.96 & 0.72 \\ 
(2)         & 0.83 & 0.68 \\ 
GS-\text{I$^{3}$} & 0.63 & 0.65 \\ 
\bottomrule
\end{tabular}
\end{table}

\begin{figure}[htbp] 
    \centering
    \includegraphics[width=0.8\linewidth]{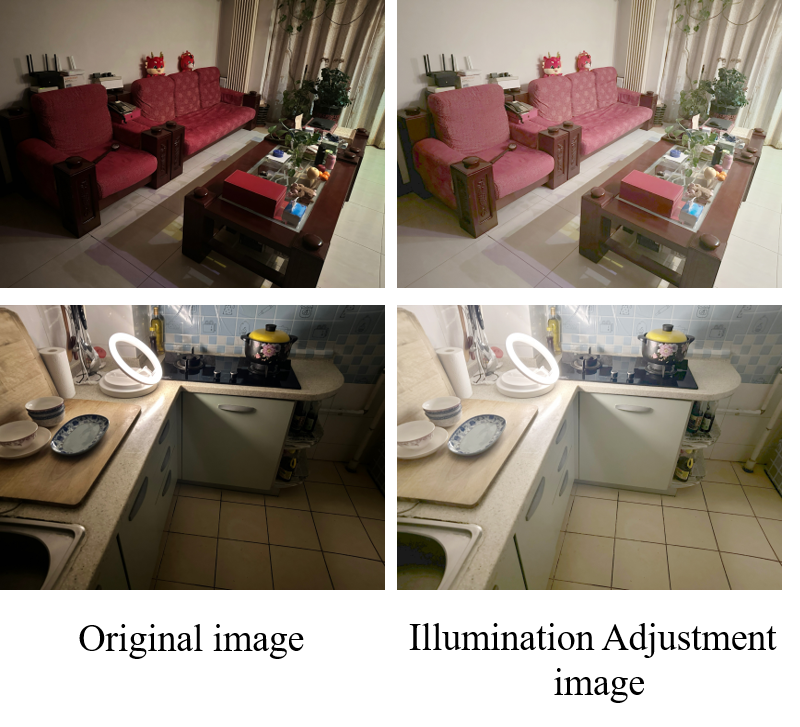}
    \caption{Comparison of different methods in the ablation study on the modified Gaussian in the dark dataset (selected scenes).}
    \label{fig:illum_img}
\end{figure}

From the quantitative results in Tab.\ref{tab:dtu_ablation}, the experimental results clearly demonstrate that using only the normal compensation has the most significant impact on the completeness and accuracy of the scene reconstruction. When only the pre-trained normal constraint is applied, the reconstruction quality is notably improved, especially under complex lighting conditions. From the qualitative results in Fig.\ref{fig:ablation_fig}, it is evident that the normal compensation corrects the direction of normals that are erroneous due to inconsistent lighting. This component effectively resolves lighting ambiguities across multiple views, enabling a more precise representation of surface geometry. It addresses issues caused by inconsistent lighting, such as surface distortion and unrealistic geometric structures.

On the other hand, as shown in Tab.\ref{tab:dtu_ablation} and Fig.\ref{fig:ablation_fig}, when only the single-view adaptive illumination adjustment equilibrium is applied, the results indicate that this method still enhances the overall reconstruction quality to some extent, particularly in handling lighting variations within individual images.This component helps to increase contrast in darker regions while preserving details in brighter areas, thereby improving the representation of surface features in these regions. As shown in Fig.\ref{fig:illum_img}, the overall brightness and contrast of the illumination adjustment image are significantly enhanced, and the objects in the image become much clearer.

From both the quantitative and qualitative results, when the two components are combined, the best reconstruction results are achieved. Futhermore, the illumination adjustment images allow the pre-trained model to predict the actual normals more accurately, reducing ambiguities caused by dark or overexposed regions. Therefore, our method leverages the synergy between these components to achieve optimal performance.

\section{Conclusion and Limitation}

 In this paper, we present GS-\text{I$^{3}$}, a novel method for accurate surface reconstruction using inconsistent illumination images. By addressing critical challenges such as exposure variation within single-view images and geometric inconsistencies across multiple views, our GS-\text{I$^{3}$} effectively reconstructs complex surfaces in challenging lighting environments. The integration of a CNN-based adaptive illumination adjustment framework and a normal compensation mechanism enables the system to adapt to complex illumination, improving both surface detail and accuracy. Extensive experimental results demonstrate the superiority of GS-\text{I$^{3}$} over existing 3DGS-based methods, particularly in terms of reducing geometric deviations and improving overall reconstruction quality. Furthermore, ablation studies confirm the efficacy of each individual component. Robustness of GS-\text{I$^{3}$} in handling lighting inconsistencies across diverse environments highlights its potential for applications in robotic exploration and 3D surface reconstruction under real-world conditions.

\bibliographystyle{IEEEtran.bst}
\bibliography{IEEEabrv,IEEEmine}

\end{document}